\documentclass[11pt,letterpaper]{article}
\usepackage{emnlp2017}
\usepackage{times}
\usepackage{latexsym}
\usepackage{multirow}
\usepackage{graphicx}
\usepackage{xcolor}
\usepackage{array}
\usepackage{bm}
\usepackage{url}
\hypersetup{draft}

\newcolumntype{H}{>{\setbox0=\hbox\bgroup}c<{\egroup}@{}}

\emnlpfinalcopy

\def\emnlppaperid{1424}

\title{Learning Language Representations for Typology Prediction}

\author{Chaitanya Malaviya \and Graham Neubig \and Patrick Littell 
  \\ {Language Technologies Institute} \\ {Carnegie Mellon University} \\
  {\tt \{cmalaviy,gneubig,pwl\}@cs.cmu.edu}}

\date{}

\begin{document}

\maketitle

\begin{abstract}
	One central mystery of neural NLP is what neural models ``know'' about their subject matter.  When a neural machine translation system learns to translate from one language to another, does it learn the syntax or semantics of the languages?  Can this knowledge be extracted from the system to fill holes in human scientific knowledge?
Existing typological databases contain relatively full feature specifications for only a few hundred languages.
Exploiting the existence of parallel texts in more than a thousand languages, we build a massive many-to-one neural machine translation (NMT) system from 1017 languages into English, and use this to predict information missing from typological databases.
Experiments show that the proposed method is able to infer not only syntactic, but also phonological and phonetic inventory features, and improves over a baseline that has access to information about the languages' geographic and phylogenetic neighbors.\footnote{Code and learned vectors are available at \url{http://github.com/chaitanyamalaviya/lang-reps}}
\end{abstract}

\section{Introduction}

Linguistic typology is the classification of human languages according to syntactic, phonological, and other classes of features, and the investigation of the relationships and correlations between these classes/features.
This study has been a scientific pursuit in its own right since the 19th century \cite{greenberg1963universals,comrie2002typology,nichols1992typology}, but recently typology has borne practical fruit within various subfields of NLP, particularly on problems involving lower-resource languages.

Typological information from sources like the World Atlas of Language Structures (WALS) \cite{wals}, has proven useful in many NLP tasks \cite{o2016survey}, such as multilingual dependency parsing \cite{ammar2016many}, generative parsing in low-resource settings~\cite{naseem2012selective,tackstrom2013target}, phonological language modeling and loanword prediction \cite{tsvetkov2016polyglot}, POS-tagging \cite{zhang2012learning}, and machine translation \cite{daiber2016reordering}.

\begin{figure}[t!]
  \centering
  \includegraphics[width=0.5\textwidth]{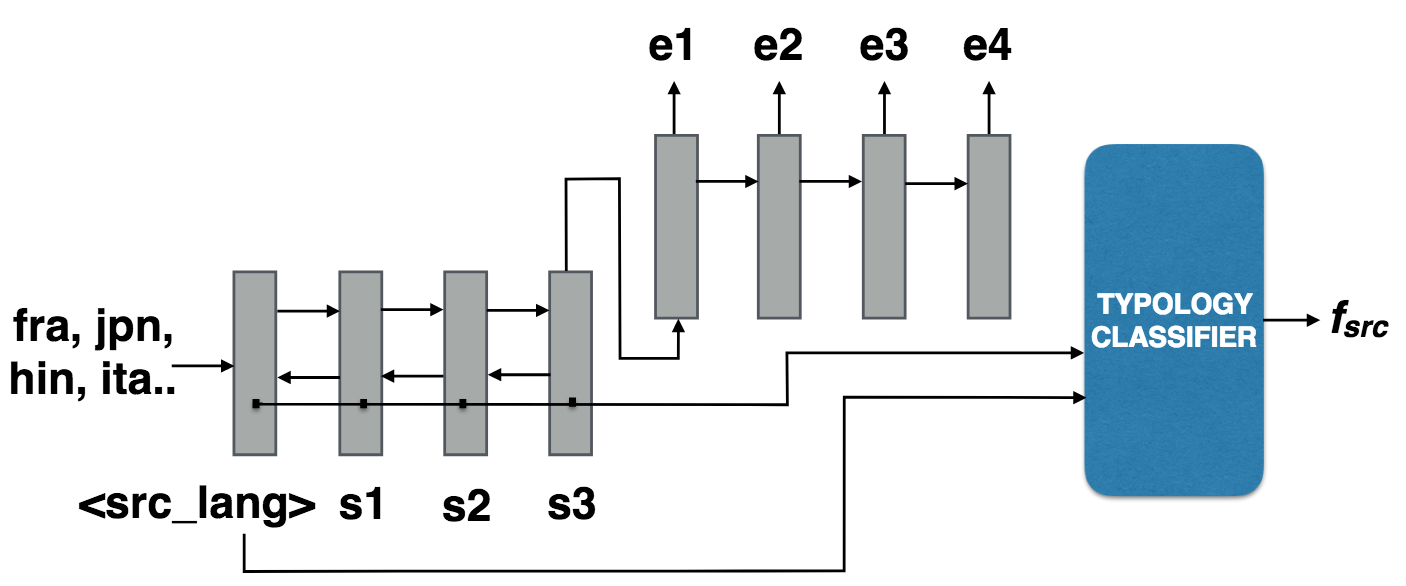}
  \caption{\label{fig:hidden_plots} Learning representations from multilingual neural MT for typology classification. (Model \textsc{MTBoth})}
\end{figure}

However, the needs of NLP tasks differ in many ways from the needs of scientific typology, and typological databases are often only sparsely populated, by necessity or by design.\footnote{For example, each chapter of WALS aims to provide a statistically balanced set of languages over language families and geographical areas, and so many languages are left out in order to maintain balance.}  In NLP, on the other hand, what is important is having a relatively full set of features for the particular group of languages you are working on.
This mismatch of needs has motivated various proposals to reconstruct missing entries, in WALS and other databases, from known entries \cite{daume2007bayesian,daume2009bayesian,coke2016classifying,littell2017uriel}.

In this study, we examine whether we can tackle the problem of inferring linguistic typology from parallel corpora, specifically by training a massively multi-lingual neural machine translation (NMT) system and using the learned representations to infer typological features for each language.
This is motivated both by prior work in linguistics \cite{bugarski1991contrastive, garcia2002contrastive} demonstrating strong links between translation studies and tools for contrastive linguistic analysis, work in inferring typology from bilingual data \cite{ostling2015word} and English as Second Language texts \cite{berzak2014reconstructing}, as well as work in NLP \cite{shi2016does,kuncoro2016recurrent,belinkov2017neural} showing that syntactic knowledge can be extracted from neural nets on the word-by-word or sentence-by-sentence level.
This work presents a more holistic analysis of whether we can discover what neural networks learn about the linguistic concepts of an entire language by aggregating their representations over a large number of the sentences in the language.


We examine several methods for discovering feature vectors for typology prediction, including those learning a language vector specifying the language while training multilingual neural language models \cite{ostling2016continuous} or neural machine translation \cite{johnson2016google} systems.
We further propose a novel method for aggregating the values of the latent state of the encoder neural network to a single vector representing the entire language.
We calculate these feature vectors using an NMT model trained on 1017 languages, and use them for typlogy prediction both on their own and in composite with feature vectors from previous work based on the genetic and geographic distance between languages \cite{littell2017uriel}.
Results show that the extracted representations do in fact allow us to learn about the typology of languages, with particular gains for syntactic features like word order and the presence of case markers.

\section{Dataset and Experimental Setup}

\paragraph{Typology Database:}
To perform our analysis, we use the URIEL language typology database \cite{littell2017uriel}, which is a collection of binary features extracted from multiple typological, phylogenetic, and geographical databases such as WALS (World Atlas of Language Structures) \cite{collins2011sswl}, PHOIBLE \cite{moran2014phoible}, Ethnologue \cite{lewis2009ethnologue}, and Glottolog \cite{glottolog}.
These features are divided into separate classes regarding syntax (e.g. whether a language has prepositions or postpositions), phonology (e.g. whether a language has complex syllabic onset clusters), and phonetic inventory (e.g. whether a language has interdental fricatives).
There are 103 syntactical features, 28 phonology features and 158 phonetic inventory features in the database.

\paragraph{Baseline Feature Vectors:}
Several previous methods take advantage of typological implicature, the fact that some typological traits correlate strongly with others, to use known features of a language to help infer other unknown features of the language \cite{daume2007bayesian,takamura2016discriminative,coke2016classifying}.
As an alternative that does not necessarily require pre-existing knowledge of the typological features in the language at hand, \newcite{littell2017uriel} have proposed a method for inferring typological features directly from the language's $k$ nearest neighbors ($k$-NN) according to geodesic distance (distance on the Earth's surface) and genetic distance (distance according to a phylogenetic family tree).
In our experiments, our baseline uses this method by taking the 3-NN for each language according to normalized geodesic+genetic distance, and calculating an average feature vector of these three neighbors.

\paragraph{Typology Prediction:}
To perform prediction, we trained a logistic regression classifier\footnote{We experimented with a non-linear classifier as well, but the logistic regression classifier performed better.} with the baseline $k$-NN feature vectors described above and the proposed NMT feature vectors described in the next section. We train individual classifiers for predicting each typological feature in a class (syntax etc). We performed 10-fold cross-validation over the URIEL database, where we train on 9/10 of the languages to predict 1/10 of the languages for 10 folds over the data.

\section{Learning Representations for Typology Prediction}
\label{sec:features}


In this section we describe three methods for learning representations for typology prediction with multilingual neural models.

\paragraph{LM Language Vector}
Several methods have been proposed to learn multilingual language models (LMs) that utilize vector representations of languages \cite{tsvetkov2016polyglot,ostling2016continuous}.
Specifically, these models train a recurrent neural network LM (RNNLM; \newcite{mikolov2010recurrent}) using long short-term memory (LSTM; \newcite{hochreiter1997long}) with an additional vector representing the current language as an input.
The expectation is that this vector will be able to capture the features of the language and improve LM accuracy.
\newcite{ostling2016continuous} noted that, intriguingly, agglomerative clustering of these language vectors results in something that looks roughly like a phylogenetic tree, but stopped short of performing typological inference.
We train this vector by appending a special token representing the source language (e.g. ``$\langle$fra$\rangle$'' for French) to the beginning of the source sentence, as shown in Fig.~\ref{fig:hidden_plots}, then using the word representation learned for this token as a representation of the language.
We will call this first set of feature vectors \textsc{LMVec}, and examine their utility for typology prediction.

\paragraph{NMT Language Vector}
In our second set of feature vectors, \textsc{MTVec}, we similarly use a language embedding vector, but instead learn a multilingual neural MT model trained to translate from many languages to English, in a similar fashion to \newcite{johnson2016google,ha2016toward}.
In contrast to \textsc{LMVec}, we hypothesize that the alignments to an identical sentence in English, the model will have a stronger signal allowing it to more accurately learn vectors that reflect the syntactic, phonetic, or semantic consistencies of various languages.
This has been demonstrated to some extent in previous work that has used specifically engineered alignment-based models \cite{lewis2008automatically,ostling2015word,coke2016classifying}, and we examine whether these results apply to neural network feature extractors and expand beyond word order and syntax to other types of typology as well.


\paragraph{NMT Encoder Mean Cell States}
Finally, we propose a new vector representation of a language (\textsc{MTCell}) that has not been investigated in previous work: the average hidden cell state of the encoder LSTM for all sentences in the language.
Inspired by previous work that has noted that the hidden cells of LSTMs can automatically capture salient and interpretable information such as syntax \cite{karpathy2015visualizing,shi2016does} or sentiment \cite{radford2017learning}, we expect that the cell states will represent features that may be linked to the typology of the language.
To create vectors for each language using LSTM hidden states, we obtain the mean of cell states ($\bm{c}$ in the standard LSTM equations) for all time steps of all sentences in each language.%
\footnote{We also tried using the mean of final hidden cell states of the encoder LSTM, but the mean cell state over all words in the sentence gave improved performance. Additionally, we tried using the hidden states $\bm{h}$, but we found that these had significantly less information and lesser variance, due to being modulated by the output gate at each time step.}

\begin{table}
\centering
\resizebox{\columnwidth}{!}{
\begin{tabular}{|c|c@{\hskip 0.1in}c|c@{\hskip 0.1in}c|c@{\hskip 0.1in}c|}
 \hline
				        & \multicolumn{2}{c|}{Syntax} & \multicolumn{2}{c|}{Phonology} & \multicolumn{2}{c|}{Inventory} \\       \cline{2-7}
 				 & -Aux           & +Aux & -Aux & +Aux & -Aux & +Aux \\
\hline \hline
\textsc{None}     & 69.91          & 83.07          & 77.92 & 86.59 & 85.17 & 90.68   \\ 
\textsc{LMVec}    & 71.32          & 82.94          & 80.80 & 86.74 & 87.51 & 89.94   \\
\textsc{MTVec}    & 74.90          & 83.31          & 82.41 & 87.64 & 89.62 & 90.94   \\
\textsc{MTCell}   & 75.91 & 85.14 & 84.33 & 88.80 & 90.01 & 90.85   \\
\textsc{MTBoth}   & \textbf{77.11}  & \textbf{86.33} & \textbf{85.77} & \textbf{89.04} & \textbf{90.06} & \textbf{91.03}  \\
\hline
\end{tabular}
}
\caption{\label{tab:res} Accuracy of syntactic, phonological, and inventory features using LM language vectors (\textsc{LMVec}), MT language vectors (\textsc{MTVec}), MT encoder cell averages (\textsc{MTCell}) or both MT feature vectors (\textsc{MTBoth}). Aux indicates auxiliary information of geodesic/genetic nearest neighbors; ``\textsc{None} -Aux'' is the majority class chance rate, while ``\textsc{None} +Aux'' is a 3-NN classification.} 
\end{table}

\section{Experiments}
\label{sec:exp}


\subsection{Multilingual Data and Training Regimen}

To train a multilingual neural machine translation system, we used a corpus of Bible translations that was obtained by scraping a massive online Bible database at \url{bible.com}.%
\footnote{A possible concern is that Bible translations may use archaic language not representative of modern usage. However, an inspection of the data did not turn up such archaisms, likely because the bulk of world Bible translation was done in the late 19th and 20th centuries. In addition, languages that do have antique Bibles are also those with many other Bible translations, so the effect of the archaisms is likely limited.}
This corpus contains data for 1017 languages.
After preprocessing the corpus, we obtained a training set of 20.6 million sentences over all languages.

The implementation of both the LM and NMT models described in \S\ref{sec:features} was done in the DyNet toolkit \cite{neubig2017dynet}.
In order to obtain a manageable shared vocabulary for all languages, we divided the data into subwords using joint byte-pair encoding of all languages \cite{sennrich2015neural} with 32K merge operations.
We used LSTM cells in a single recurrent layer with 512-dimensional hidden state and input embedding size. The Adam optimizer was used with a learning rate of 0.001 and a dropout of 0.5 was enforced during training.


\subsection{Results and Discussion}
\label{sec:res_disc}

The results of the experiments can be found in Tab.~\ref{tab:res}.
First, focusing on the ``-Aux'' results, we can see that all feature vectors obtained by the neural models improve over the chance rate, demonstrating that indeed it is possible to extract information about linguistic typology from unsupervised neural models.
Comparing \textsc{LMVec} to \textsc{MTVec}, we can see a convincing improvement of 2-3\% across the board, indicating that the use of bilingual information does indeed provide a stronger signal, allowing the network to extract more salient features.
Next, we can see that \textsc{MTCell} further outperforms \textsc{MTVec}, indicating that the proposed method of investigating the hidden cell dynamics is more effective than using a statically learned language vector. Finally, combining both feature vectors as \textsc{MTBoth} leads to further improvements. To measure statistical significance of the results, we performed a paired bootstrap test to measure the gain between \textsc{None+Aux} and \textsc{MTBoth+Aux} and found that the gains for syntax and inventory were significant (p=0.05), but phonology was not, perhaps because the number of phonological features was fewer than the other classes (only 28).

When further using the geodesic/genetic distance neighbor feature vectors, we can see that these trends largely hold although gains are much smaller, indicating that the proposed method is still useful in the case where we have a-priori knowledge about the environment in which the language exists.
It should be noted, however, that the gains of \textsc{LMVec} evaporate, indicating that access to aligned data may be essential when inferring the typology of a new language. We also noted that the accuracies of certain features decreased from \textsc{None-Aux} to \textsc{MTBoth-Aux}, particularly gender markers, case suffix and negative affix, but these decreases were to a lesser extent in magnitude than the improvements.

\begin{table}
\small
\centering
\resizebox{\columnwidth}{!}{
\begin{tabular}{|c|r@{\hskip 0.1in}r@{\hskip 0.1in}r|}
\hline \textbf{Feature} & \textbf{\textsc{None}} & \multicolumn{1}{c}{\textbf{\textsc{MT}}} & \textbf{Gain} \\ \hline \hline
S\_NUMERAL\_AFTER\_NOUN & 37.40 & 81.26 & 43.86 \\
S\_NUMERAL\_BEFORE\_NOUN & 46.49 & 83.22 & 36.73 \\
S\_POSSESSOR\_AFTER\_NOUN & 42.05 & 75.60 & 33.55 \\
S\_OBJECT\_BEFORE\_VERB & 50.97 & 80.89 & 29.92 \\
S\_ADPOSITION\_AFTER\_NOUN & 52.41 & 79.10 & 26.69 \\ \hline
P\_UVULAR\_CONTINUANTS & 77.57 & 97.37 & 19.80 \\
P\_LATERALS & 67.30 & 86.48 & 19.18 \\
P\_LATERAL\_L & 64.05 & 78.16 & 14.10 \\
P\_LABIAL\_VELARS & 82.16 & 95.93 & 13.76 \\
P\_VELAR\_NASAL\_INITIAL & 72.14 & 85.82 & 13.68 \\ \hline
I\_VELAR\_NASAL & 39.89 & 62.08 & 22.20 \\
I\_ALVEOLAR\_LATERAL\_APPROXIMANT & 60.92 & 79.32 & 18.40 \\
I\_ALVEOLAR\_NASAL & 81.49 & 92.98 & 11.48 \\
I\_VOICED\_LABIODENTAL\_FRICATIVE & 65.75 & 77.10 & 11.36 \\
I\_VOICELESS\_PALATAL\_FRICATIVE & 82.41 & 93.66 & 11.25 \\ \hline
\hline
\end{tabular}
}
\caption{\label{tab:feat-table} Top 5 improvements from ``\textsc{None} -Aux'' to ``\textsc{MTBoth} -Aux'' in the syntax (``S\_''), phonology (``P\_''), and inventory (``I\_'') classes.}
\end{table}

\begin{figure}[t!]
  \centering
  \includegraphics[width=0.5\textwidth]{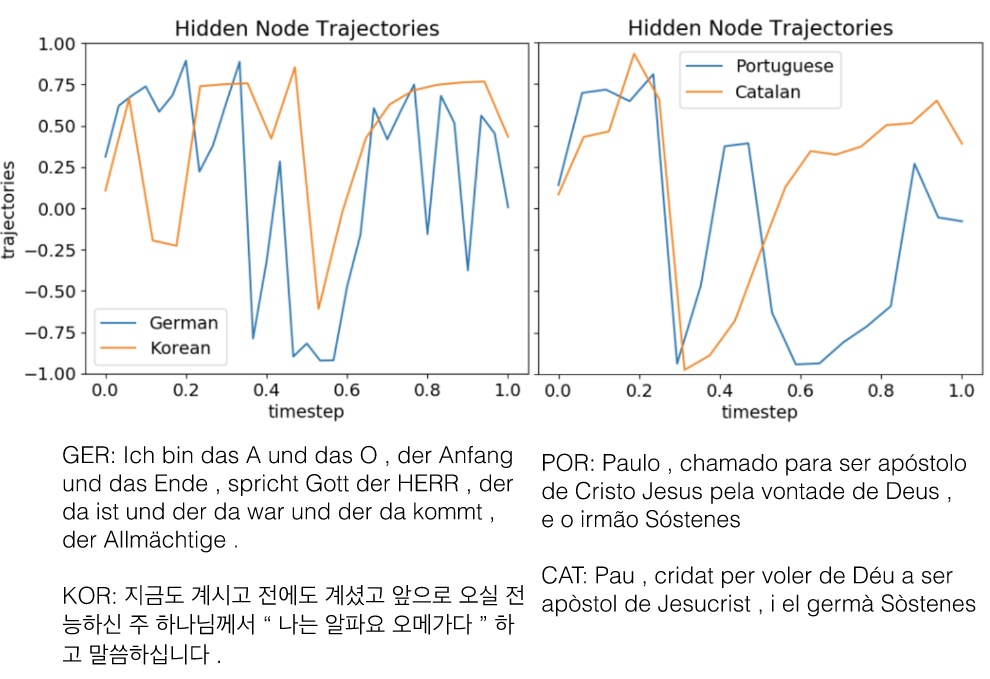}
  \caption{\label{fig:node_traj} Cell trajectories for sentences in languages where  S\_OBJ\_BEFORE\_VERB is either active or inactive.}
\end{figure}

Interestingly, and in contrast to previous methods for inferring typology from raw text, which have been specifically designed for inducing word order or other syntactic features \cite{lewis2008automatically,ostling2015word,coke2016classifying}, our proposed method is also able to infer information about phonological or phonetic inventory features.
This may seem surprising or even counter-intuitive, but a look at the most-improved phonology/inventory features (Tab.~\ref{tab:feat-table}) shows a number of features in which languages with the ``non-default" option (e.g. having uvular consonants or initial velar nasals, \emph{not} having lateral consonants, etc.) are concentrated in particular geographical regions.  For example, uvular consonants are not common world-wide, but are common in particular geographic regions like the North American Pacific Northwest and the Caucasus \cite{wals-6}, while initial velar nasals are common in Southeast Asia \cite{wals-9}, and lateral consonants are \emph{uncommon} in the Amazon Basin \cite{wals-8}.  Since these are also regions with a particular and sometimes distinct syntactic character, we think the model may be finding regional clusters through syntax, and seeing an improvement in regionally-distinctive phonology/inventory features as a side effect.

Finally, given that \textsc{MTCell} uses the feature vectors of the latent cell state to predict typology, it is of interest to observe how these latent cells behave for typologically different languages.
In Fig.~\ref{fig:node_traj} we examine the node that contributed most to the prediction of ``S\_OBJ\_BEFORE\_VERB'' (the node with maximum weight in the classifier) for German and Korean, where the feature is active, and Portuguese and Catalan, where the feature is inactive.
We can see that the node trajectories closely track each other (particularly at the beginning of the sentence) for Portuguese and Catalan, and in general the languages where objects precede verbs have higher average values, which would be expressed by our mean cell state features. The similar trends for languages that share the value for a typological feature (S\_OBJ\_BEFORE\_VERB) indicate that information stored in the selected hidden node is consistent across languages with similar structures.







\section{Conclusion and Future Work}

Through this study, we have shown that neural models can learn a range of linguistic concepts, and may be used to impute missing features in typological databases. In particular, we have demonstrated the utility of learning representations with parallel text, and results hinted at the importance of modeling the dynamics of the representation as models process sentences. We hope that this study will encourage additional use of typological features in downstream NLP tasks, and inspire further techniques for missing knowledge prediction in under-documented languages.


\section*{Acknowledgments}

We thank Lori Levin and David Mortensen for their useful comments and also thank the reviewers for their feedback about this work.

\bibliography{emnlp2017}
\bibliographystyle{emnlp_natbib}

\end{document}